\documentclass[a4paper,preprint, pre,showpacs,footinbib,superscriptaddress]{revtex4-1}
\usepackage{amsmath,amssymb}
\usepackage{makeidx}
\usepackage{amsfonts}
\usepackage[ansinew]{inputenc}
\usepackage[usenames,dvipsnames]{pstricks}
\usepackage{epsfig,graphicx}
\usepackage{color}

\usepackage{amsthm}

\newcommand{\bb}[1]{\mathbf{\boldsymbol{#1}}}
\newcommand{\s}{\bb{s}}
\renewcommand{\S}{\bb{S}}

\newcommand{\y}{\bb{y}}

\newcommand{\btheta}{\bb{\theta}}
\newcommand{\bpi}{\bb{\pi}}

\newcommand{\eq}[1]{\begin{align}#1\end{align}}{\ignorespacesafterend}
{\ignorespacesafterend}

\makeindex
\begin{document}

\title{Learning Mixtures of Ising Models using Pseudolikelihood } 
\author{Onur Dikmen}
\affiliation{Department of Information and Computer Science, University of Helsinki, P.O. Box 68, FI-00014,  Finland.}
\date{\today}

\begin{abstract}
Maximum pseudolikelihood method has been among the most important methods for learning parameters of statistical physics models, such as Ising models. In this paper, we study how pseudolikelihood can be derived for learning parameters of a mixture of Ising models. The performance of the proposed approach is demonstrated for Ising and Potts models on both synthetic and real data.
\end{abstract}

\maketitle

Inference in models of statistical physics, such as Ising models, is generally challenging because of the intractable partition function (normalizing constant). Inference could be in the form of  obtaining statistical properties of the model, e.g., magnetization, (forward problem) or learning parameters of the model which are most likely to have generated a given dataset (inverse problem). Both forms of the problem proves difficult in the presence of unknown partition functions. Methods based on Markov chain Monte Carlo (MCMC)~\cite{ackley85learning}  or mean field approaches~\cite{opper2001advanced} have been widely used for both problems. Maximum pseudolikelihood (MPL)~\cite{besag75statistical}  is another popular method successfully applied to many inverse problem applications~\cite{ekeberg13improved,aurell12inverse}. MPL is a consistent estimator for parameters, i.e., asymptotically (number samples going to infinity) it recovers true parameters~\cite{gidas1988consistency}. Therefore, higher accuracy is expected especially  when high amount of data is available. For example, in the field of direct coupling analysis, MPL is currently known as the most accurate method~\cite{ekeberg13improved,Morcos2011direct}.

In this paper, we study the problem of learning parameters of a mixture of Ising models. It may be the case that data at hand can be explained / may have been generated by more than one Ising model. Learning one set of parameters (one model) in this case may cause some (or even none) of the data samples not to be represented by the model. Below, we propose a method based on pseudolikelihood to learn parameters of a mixture model. Training a mixture of $K$ Ising models does not have much overhead and is as efficient as training $K$ separate Ising models on the same data.

A mixture of Ising models is a superposition of $K$ Ising models and its pdf is given by
\eq{
\label{eqn:marg}
p(\s|\bpi,\btheta) = \sum_{k=1}^K \pi_k p_k(\s|\btheta_k)\,,
}
where $\pi_k$ are mixing coefficients with $\sum_{k=1}^K\pi_k=1$ and $\btheta_k$ are parameters of $k$th Ising model which comprises coupling parameters $J_{ij}^k$, external fields $h_i^k$ and inverse temperature $\beta^k=1/T^k$. The density of an individual Ising model is
\eq{
p_k(\s|\btheta_k) = \frac{\phi_k(\s;\btheta_k)} {Z_k(\btheta_k)} = \frac{\exp(\beta^k \sum_i h_i^k s_i + \beta^k \sum_{i < j} J_{ij}^k s_i s_j)} {Z_k(\btheta_k)}\,.
}
where $Z_k(\btheta_k)$ is the partition function (normalizing constant) of which computation is intractable. The observed variables $\s$ are binary of dimension $N$, i.e., $s\in\{-1,+1\}^N$.

In the inverse problem of mixture of Ising models, the goal is to learn parameters of the model $\{\pi_k, \btheta_k\}_{k=1}^K$ form a dataset $\S=\{\s_b\}_{b=1}^B = \{s_{bn}\}_{b=1,n=1}^{B,N}$ with $B$ samples. When $Z_k(\btheta_k)$ are analytically available, the standard way for the inverse problem is maximum likelihood estimation
\eq{
\{\pi_k^*, \btheta_k^*\}_{k=1}^K\ = \arg\max_{\{\pi_k, \btheta_k\}_{k=1}^K} \log p(\S|\bpi,\btheta) = \arg\max_{\{\pi_k, \btheta_k\}_{k=1}^K} \sum_b \log \sum_{k=1}^K \pi_k p_k(\s_b|\btheta_k)\,.
}

If we maximize $\log p(\S|\bpi,\btheta)$ w.r.t mixing coefficients $\pi_k$ (using a Lagrange multiplier for $\sum \pi_k=1$), we obtain
\eq{
\label{eqn:gamma}
\pi_k \leftarrow \sum_{b=1}^B \frac{\pi_k p_k(\s_b|\btheta_k)}{\sum_{j=1}^K \pi_j p_j(\s_b|\btheta_j)}\equiv \sum_{b=1}^B\gamma_{bk}\,,
}
where we may dub the right hand side ``responsibility" of mixture $k$ for sample $b$ and represent it with $\gamma_{bk}$. Of course, $\gamma_{bk}$ is not analytically available due to unknown normalizing constants, $Z_k$.

$\gamma_{bk}$ can be approximated however by approximating $Z_k$ using MCMC methods or mean field lower bounds. Then, the gradient of the log likelihood w.r.t $\btheta_k$ is written
\eq{
\frac{\partial \log p(\S|\bpi,\btheta)}{\partial \btheta_k} = \sum_{b=1}^B \gamma_k \left(\frac{\partial\log\phi_k(\s_b;\btheta_k)}{\partial\btheta_k} - \frac{\partial\log Z_k(\btheta_k)}{\partial\btheta_k}\right)\,.
}
This can be accomplished by again using MCMC estimates of $\partial\log Z_k(\btheta_k)/\partial\btheta_k$, Moreover, it gives a hint that other standard methods like pseudolikelihood or mean field equations can be derived for optimization of $\btheta_k$'s from individual Ising models. Below, we derive pseudolikelihood for optimization of MoI.


We introduce the latent variable $\y_b$ for each data sample. $\y_b$ has a multinomial distribution with one trial, i.e., $y_{bk}\in \{0,1\}$, $\sum_k y_{bk}=1$, $p(y_{bk}=1)=\pi_k$, $p(\y_b)=\prod_k \pi_k^{y_{bk}}$. $\y_b$ determines which mixture the data point $\s_b$ belongs to.

Each mixture is given by an Ising model based on the following conditional probability
\eq{
p(\s_b|y_{bk}=1) = p_k(\s_b|\btheta_k)\,,
}
which in turn leads to
\eq{
p(\s_b|\y_b,\btheta) = \prod_k p_k(\s_b|\btheta_k)^{y_{bk}}\,.
}
The marginal of this model is the same as in \eqref{eqn:marg} and the posterior is $p(y_{bk}=1|\s_b)=\gamma_{bk}$ as in \eqref{eqn:gamma}.

In order to obtain the full conditional distributions of the observed variables, we make use of the following equality
\eq{
\label{eqn:u_n}
\frac{p(\s_{b,-n}|\btheta)}{p(\s_b|\btheta)}=\sum_{\y_b} \frac{p(\s_{b,-n},\y_b|\btheta)}{p(\s_b,\y_b|\btheta)}p(\y_b|\s_b,\btheta)\equiv\frac{1}{u_{bn}}\,,
}
where $\s_{b,-n}$ denotes the vector $\s_b$ with $n$th variable flipped. With $1/u_{bn}$ we can write log pseudolikelihood as follows
\eq{
\log\mathrm{PL} = \frac{1}{B} \sum_b \sum_n \log(u_{bn}/(1+u_{bn}))\,.
}

For the mixture Ising model, \eqref{eqn:u_n} is simplified as follows
\eq{
1/u_{bn} &= \sum_{\y_b} \prod_k \frac{\phi_k(\s_{b,-n}|\btheta_k)^{y_{bk}}}{\phi_k(\s_b|\btheta_k)^{y_{bk}}} p(\y_b|\s_b,\bpi,\btheta)\\
&= \sum_k \frac{\phi_k(\s_{b,-n}|\btheta_k)}{\phi_k(\s_b|\btheta_k)} p(y_{bk}=1|\s_b,\bpi,\btheta)\,.
}
This suggests that we can build an EM-like iterative algorithm where we first estimate the responsibilities at iteration $t$, $p(y_{bk}=1|\s_b,\bpi^t,\btheta^t)$, then update mixing coefficients $\bpi$ and the parameters of individual Ising models $\btheta_k$. Responsibilities can be estimated using any of the approaches mentioned above, such as MCMC, mean field lower bound, etc. In this work, we propose to estimate them using pseudolikelihood, i.e., use $\mathrm{PL}_k$ instead of $p_k$ in \eqref{eqn:gamma}. This results in a very efficient method where both estimation and optimization steps are done using pseudolikelihood.

\begin{figure}[htb]
\begin{center}
\begin{tabular}{cc}
  Individual Datasets:&\\
  \epsfig{figure=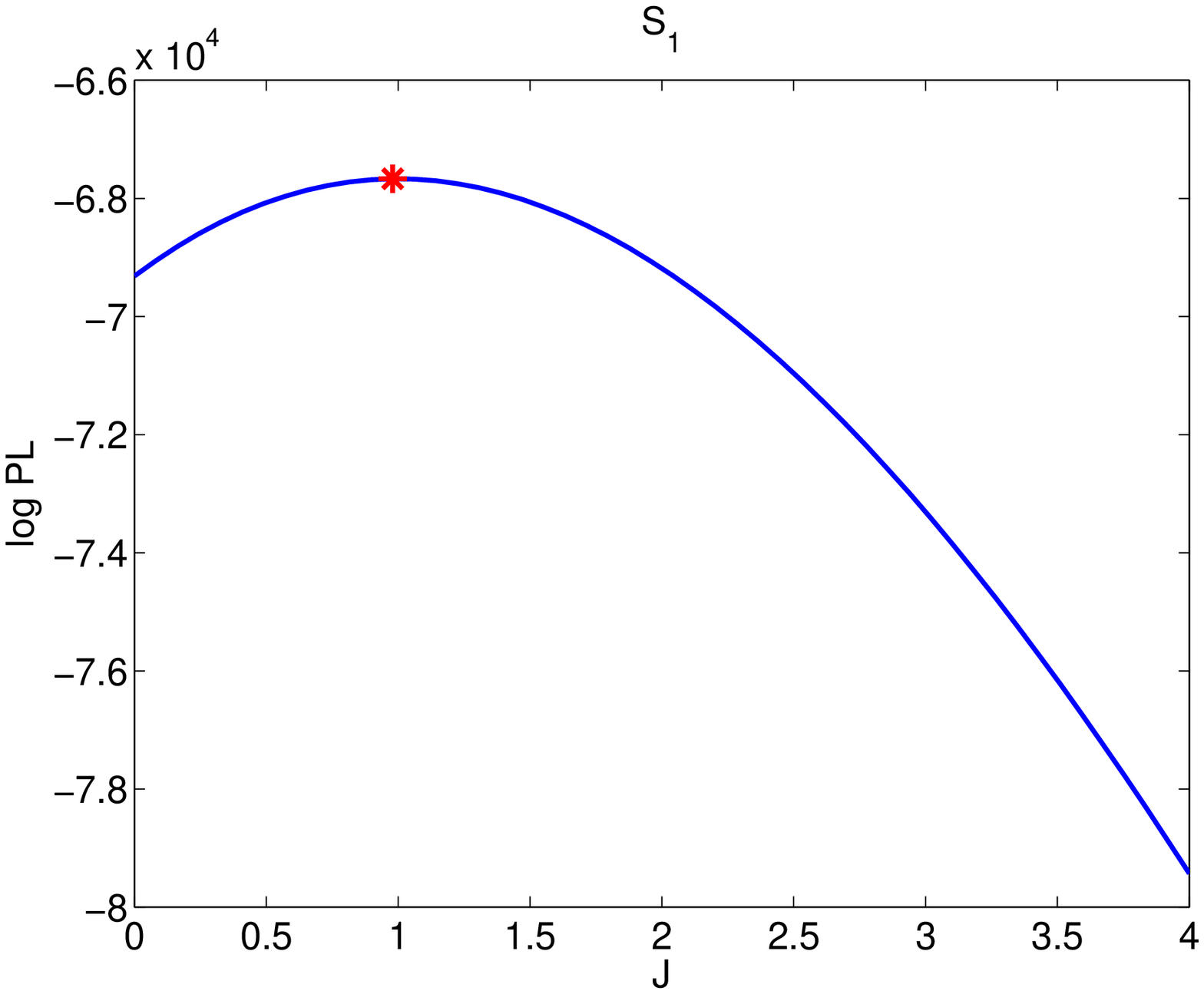,width=6cm}&
  \epsfig{figure=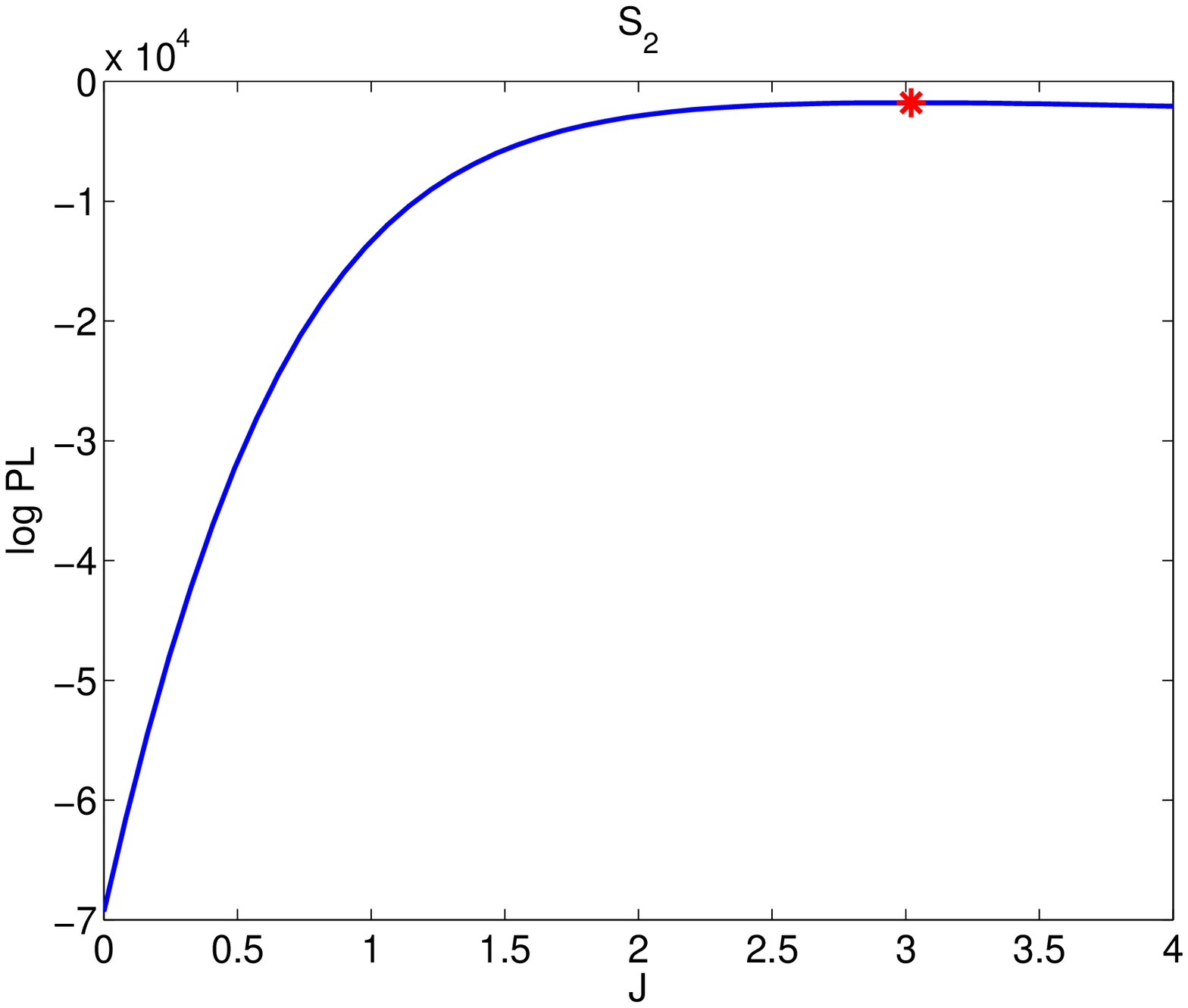,width=6cm}\\
  Concatenated Dataset: & Single Ising (left), Mixture of Isings (right)\\
  \epsfig{figure=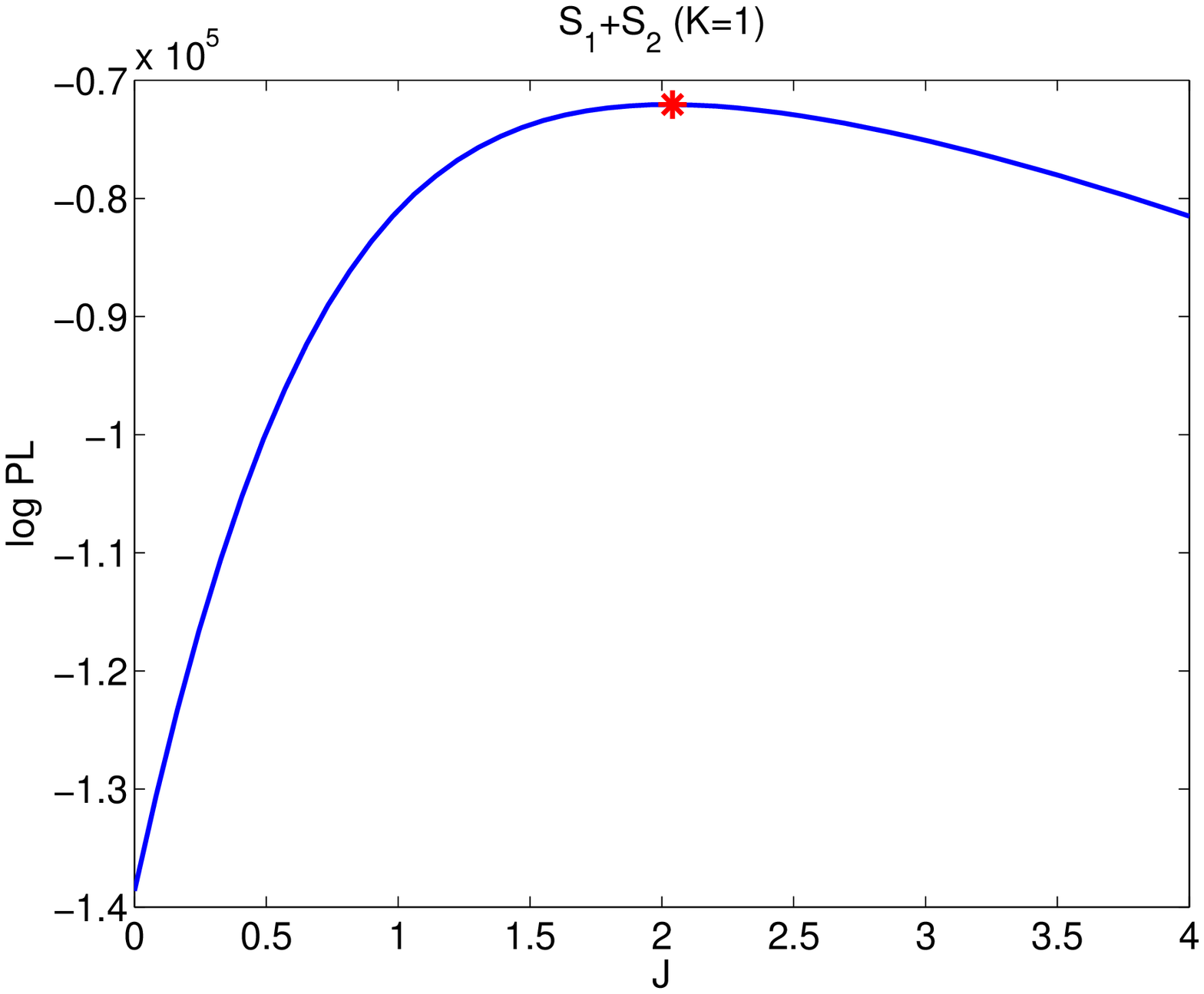,width=6cm}&
  \epsfig{figure=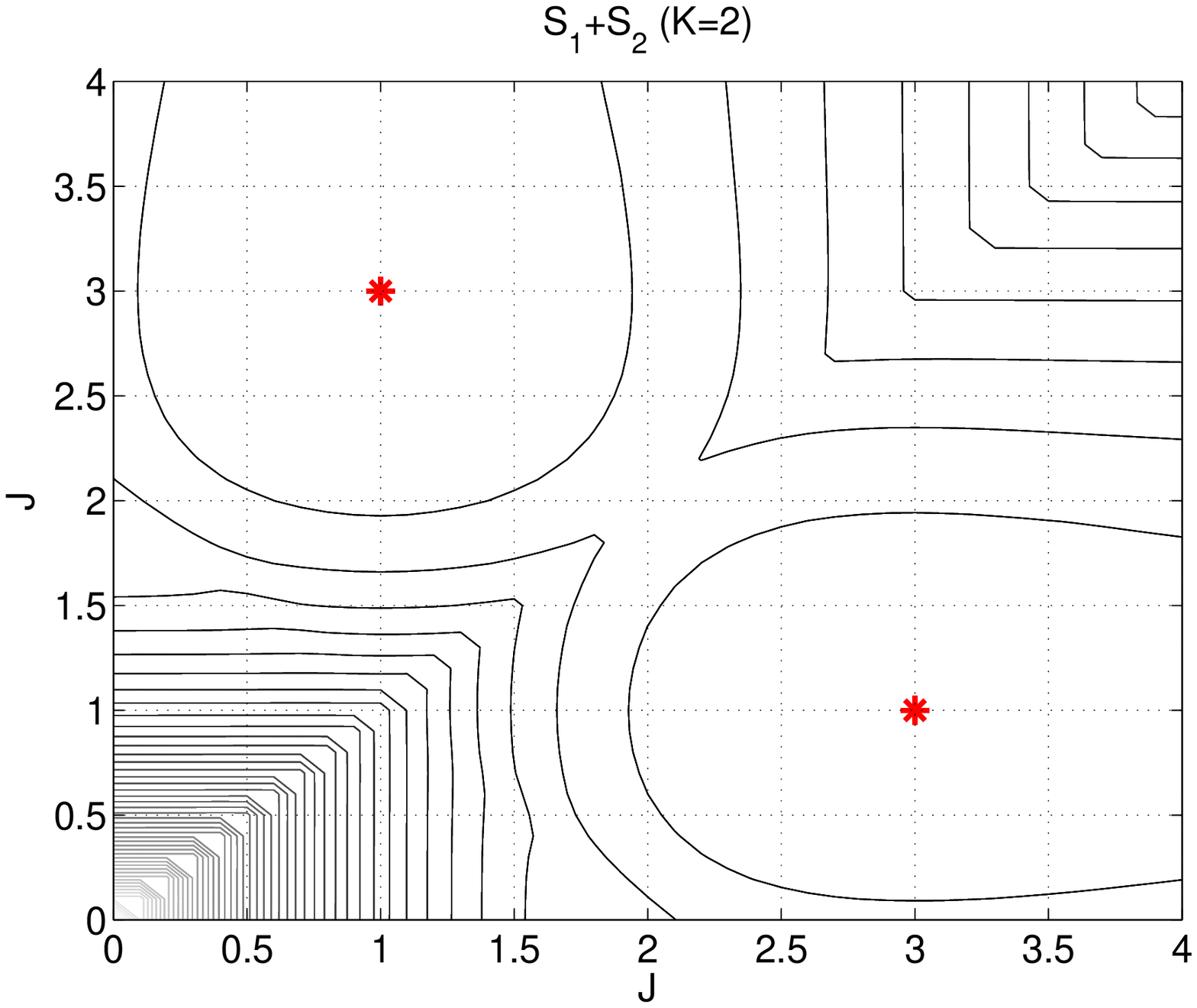,width=6cm}
\end{tabular}
\end{center}
\caption{Infinite range model. $N=1000$, $B=100$, $\beta=0.001$.}
\label{fig:ir_model}
\end{figure}

We used infinite range (IR) Ising model to show the good performance of pseudolikelihood on both single models and mixtures. IR models have the same coupling parameter between all pairs of variables, i.e., $J_{ij} = J$. We fixed the external fields to zero, $h_i = 0$ for all variables and $\beta=0.001$. Then, generated two datasets $\S_1$ and $\S_2$ using $J=1$ and $J=3$. Pseudolikelihood surfaces w.r.t. $J$ on these two datasets are given in the top row of Fig.\ref{fig:ir_model}. The $J^*$ values which maximize these pseudolikelihood curves agree with the parameter values used to generate data. If we concatenate the datasets, i.e., $\S=\S_1+\S_2$, and obtain the pseudolikelihood curve on this single dataset (Fig.\ref{fig:ir_model} bottom row, left plot), we see that the optimal $J$ value is neither of the original values (1 or 3). On the contrary, if we model $\S$ with a MoI with two mixtures and plot the pseudolikelihood surface (Fig.\ref{fig:ir_model} bottom row, right plot), we see that the optimal parameter values  $J^*=1$ and $J^*=3$ coincide with the original values. Note that, the symmetry in the model results in unidentifiability, i.e., there are two symmetric peaks.

\begin{figure}[htb]
\begin{center}
\begin{tabular}{cc}
  Ground Truth:&\\
  \epsfig{figure=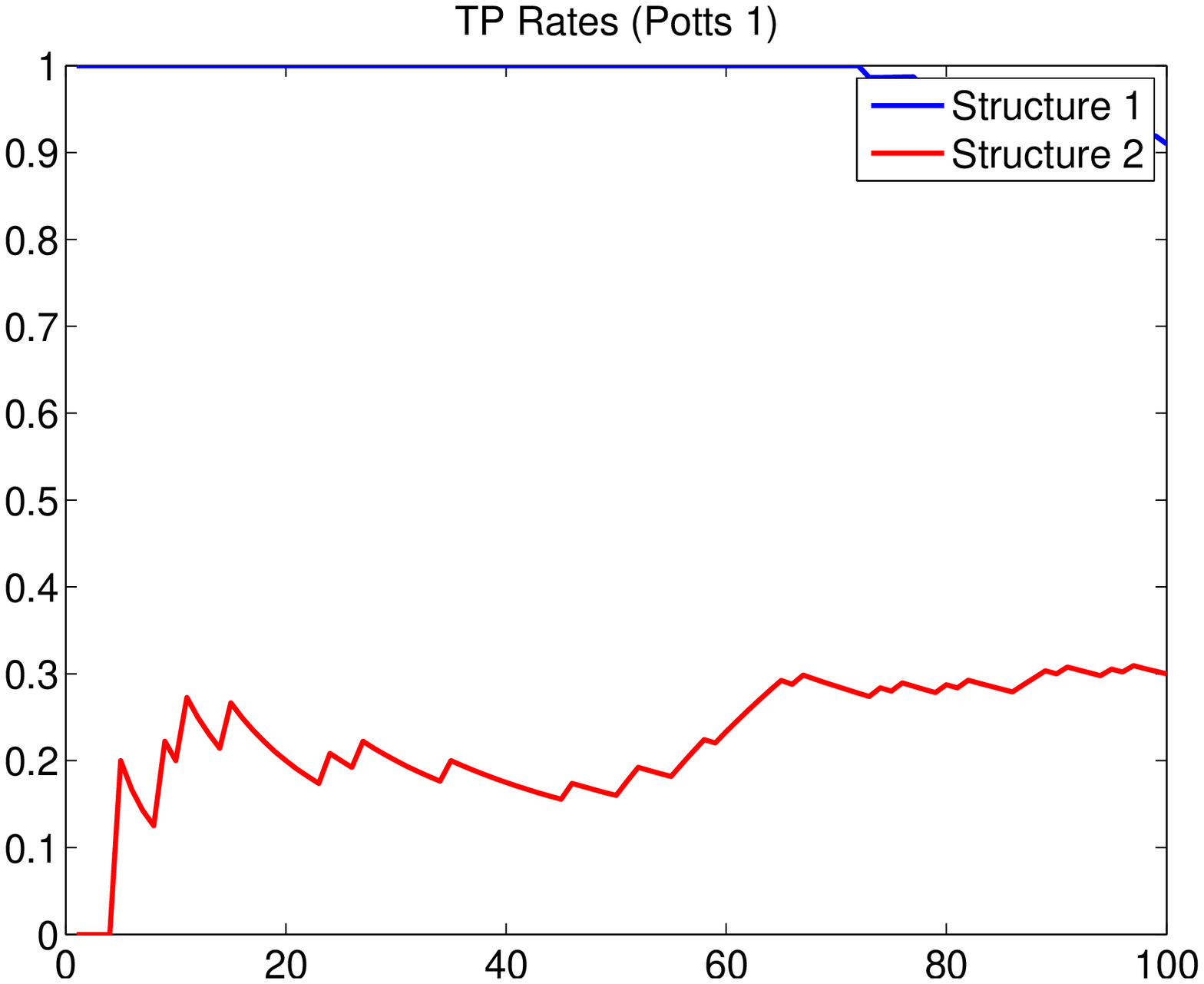,width=5.8cm}&
  \epsfig{figure=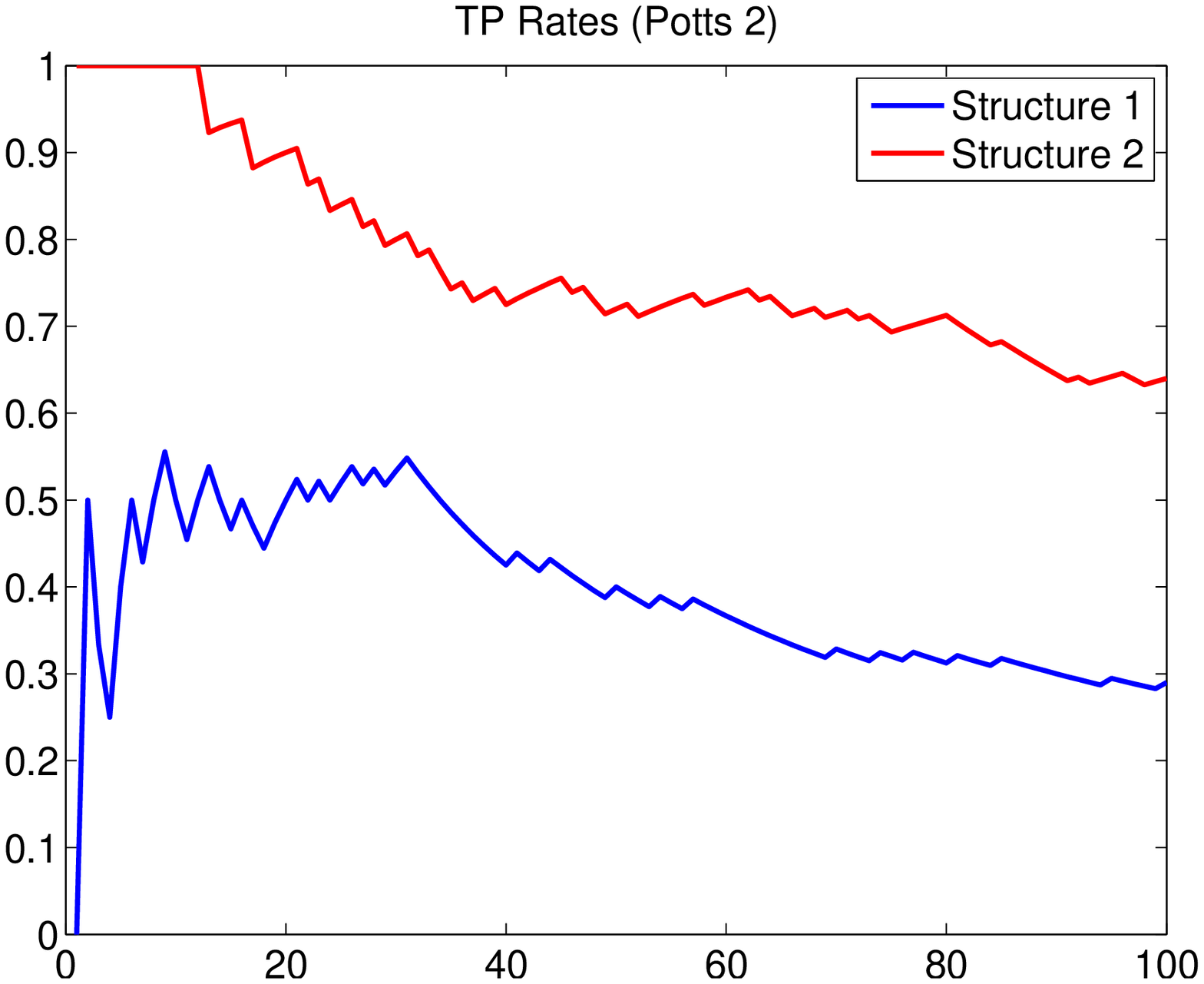,width=5.8cm}\\
  Random Initialization:&\\
  \epsfig{figure=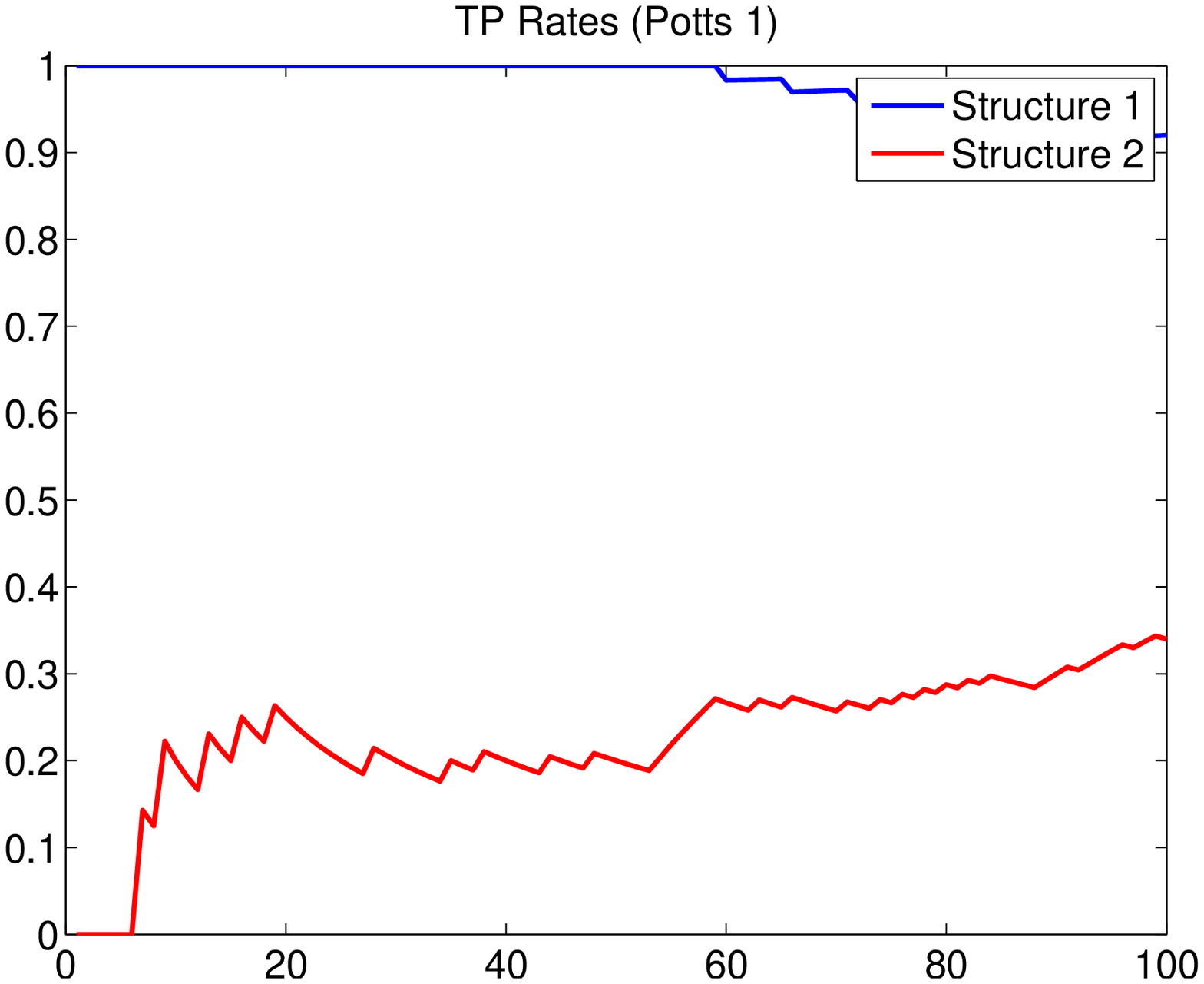,width=5.8cm}&
  \epsfig{figure=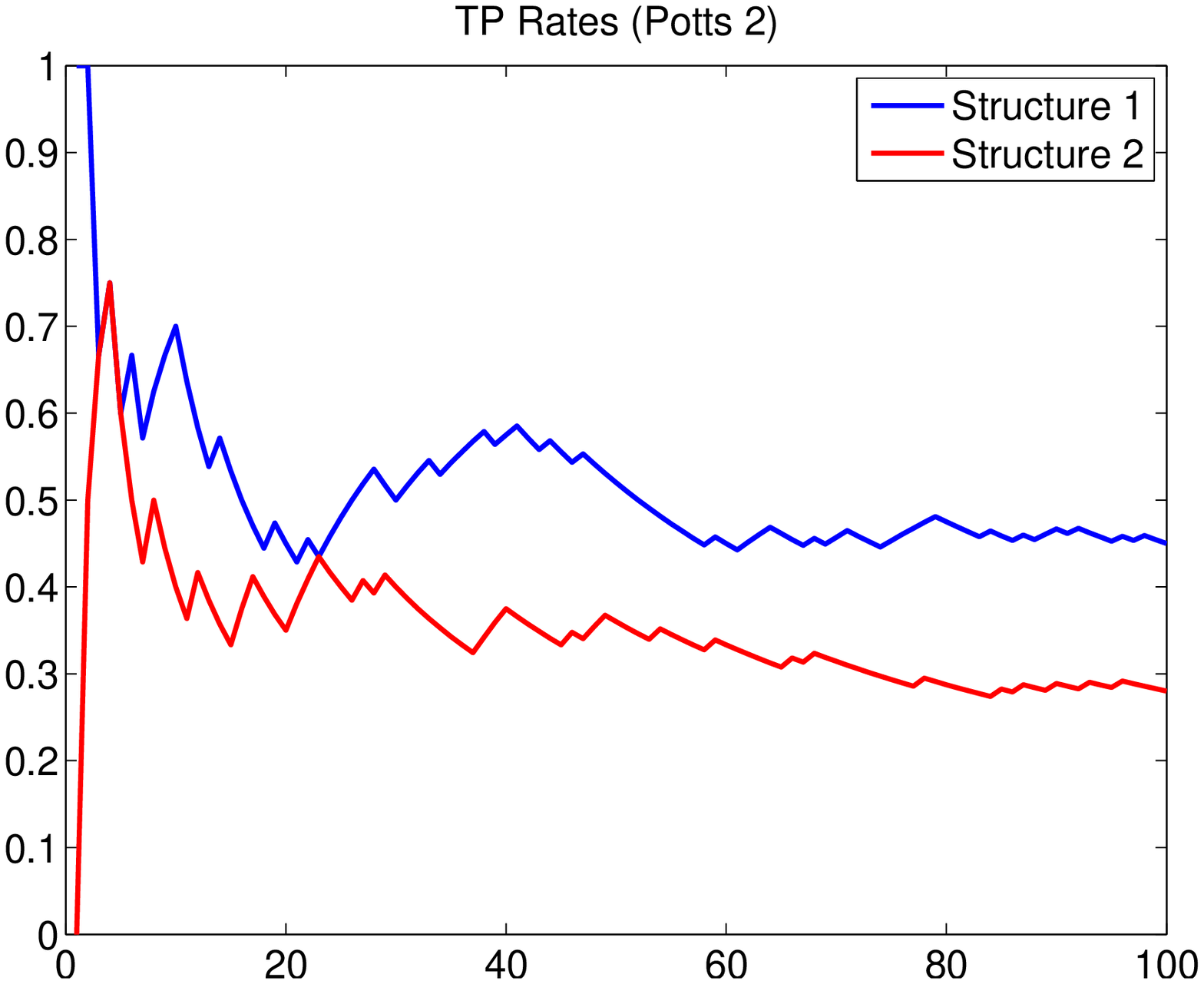,width=5.8cm}\\
  Better Initialization:&\\
  \epsfig{figure=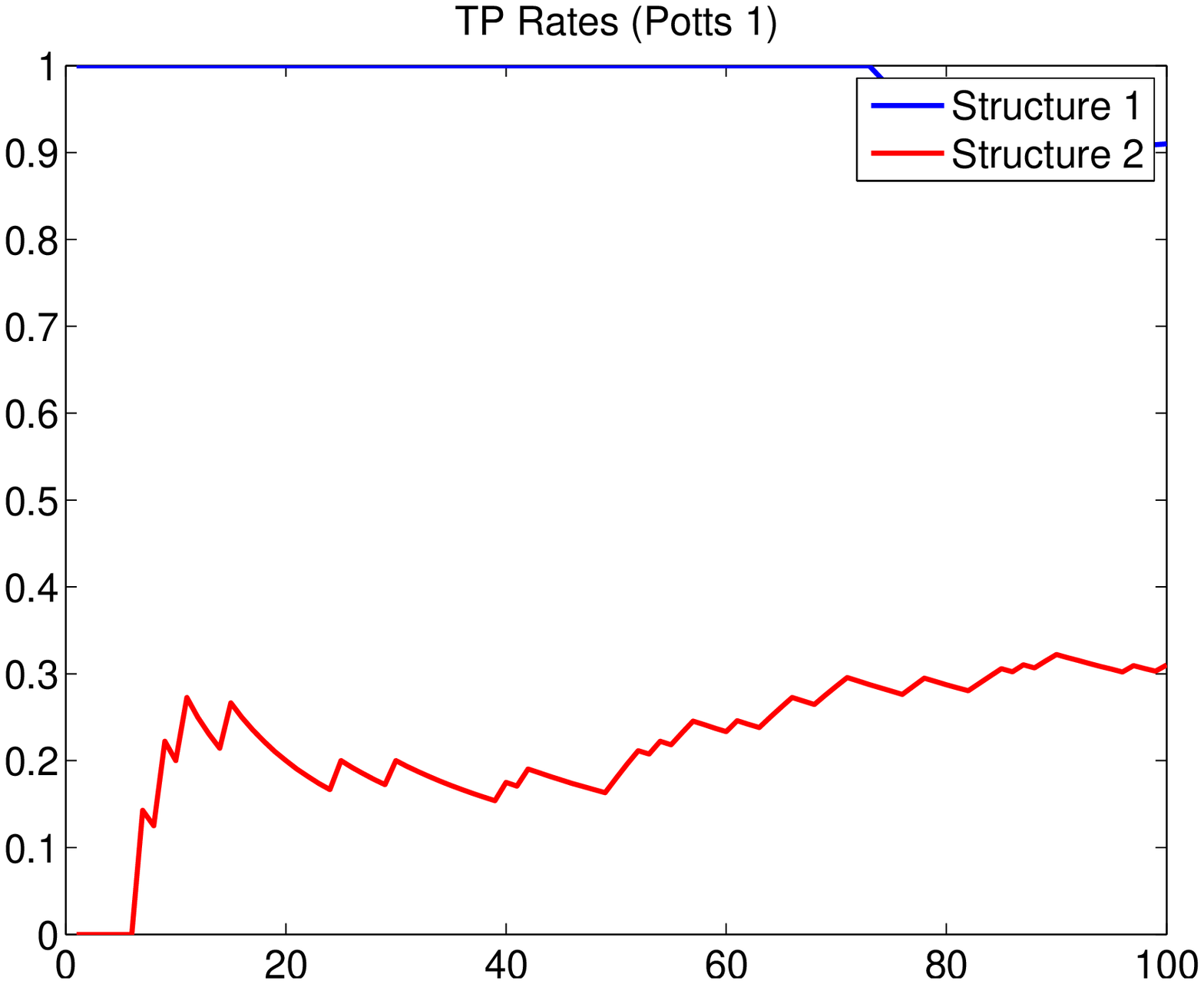,width=5.8cm}&
  \epsfig{figure=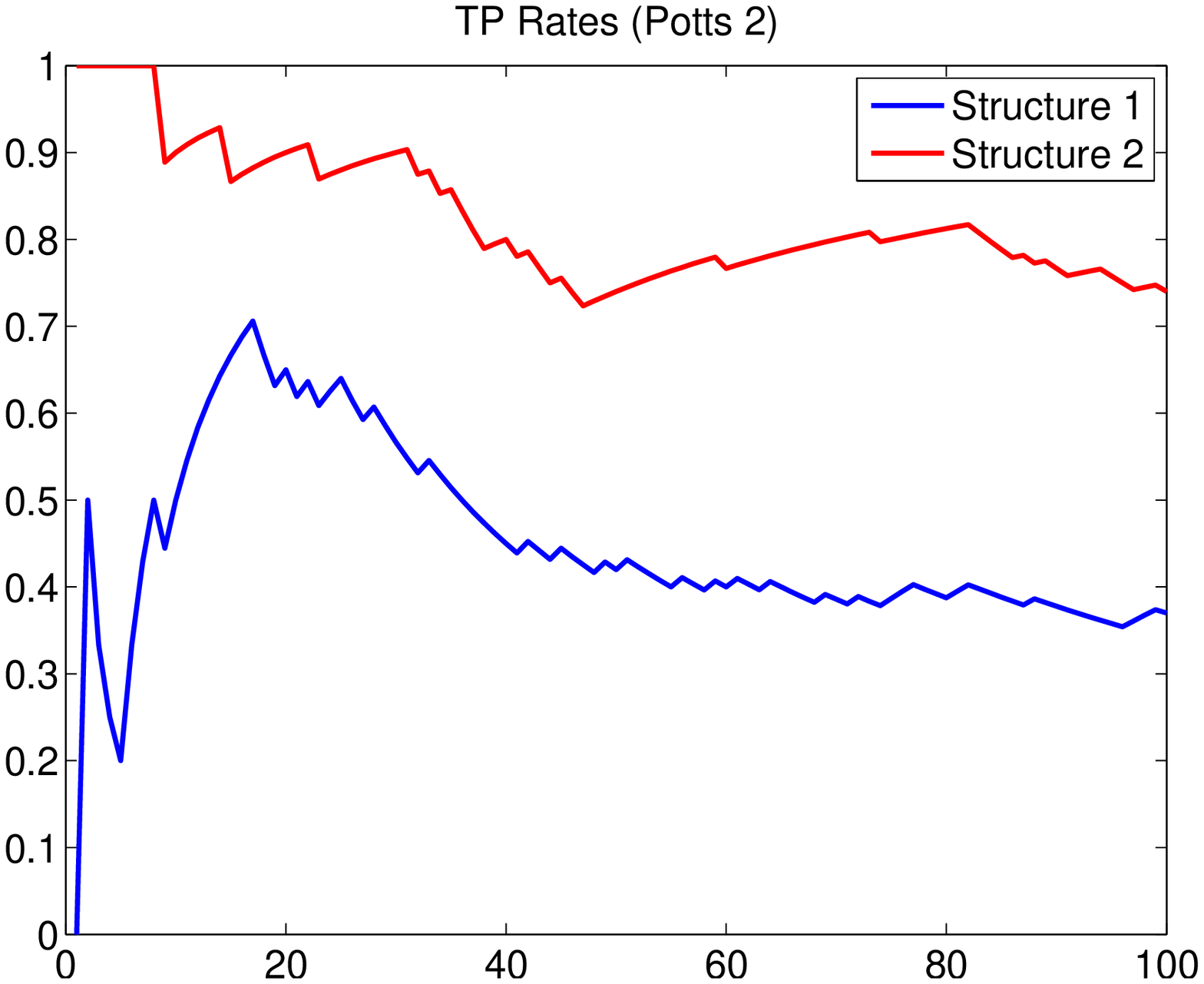,width=5.8cm}
\end{tabular}
\end{center}
\caption{PF00076 \& PF00105 concatenated. TP rates on individual families (top row): PF00076 (left), PF00105 (right); MoP results with random initialization (middle row) and with a more informed initialization (bottom row).}
\label{fig:76_105}
\end{figure}

The proposed MoI learning method can also be extended to mixtures of Potts (MoP) models. A Potts model is basically an Ising model where variables are discrete with more than two states. Pseudolikelihood for MoP is easily accomplished by considering the flipped state $\s_{b,-n}$ as a collection of all states except $\s_{bn}$. We tested PL for MoP in direct coupling analysis for protein structure prediction problem where PL is one of the most successful approaches~\cite{ekeberg13improved}. We consider two protein families (PF00076 and PF00105) from the PFAM database which has the same aligned sequence length ($N=70$). Contact-detection results on individual datasets with single Potts models are presented in Fig.~\ref{fig:76_105} (top row). The plots are true positive (TP) rates w.r.t. the number of predicted contacts, based on pairs with $|i-j|>4$. 

We concatenated these two datasets and tried to learn two different sets of parameters using PL for MoP with $K=2$. With random initialization of $\gamma_{bk}$'s (Fig.~\ref{fig:76_105}, middle row), the TP rates obtained with one of the Potts models (right) is not as good as the single Potts model on PF00105 only. However, the results can be improved substantially with a more informed initialization. This can be achieved in many ways. We considered here choosing $K$ samples (codewords) furthest away from each other from a random subset and assigning the rest of the samples to the closest codeword through $\gamma_{bk}$. With this better initialization, the TP rates are as good as (Fig.~\ref{fig:76_105}, bottom row) the single Potts models on original (separate) datasets.

In summary, we proposed a very simple but efficient way of training mixtures of Ising models. We showed that when the data includes samples with different characteristics, indeed a mixture model explains the data better and the overall inference is beneficial. Our method of choice here was pseudolikelihood, but a similar approach can be pursued using mean field as well. The method is  as efficient as learning $K$ Ising models, without bringing an overhead stemming from handling mixtures. An immediate future work is model selection, i.e., selecting the optimal $K$ also from data.

\begin{acknowledgments} The work of Onur Dikmen (OD) is supported by the Academy of Finland (Finnish Centre of Excellence in Computational Inference Research COIN, 251170). OD would like to thank Erik Aurell and Marcin J. Skwark for introducing the problem and kindly sharing their expertise in interpreting the results.
 \end{acknowledgments}

\bibliography{moi}
\bibliographystyle{apsrev4-1}
\end{document}